# Multilevel Thresholding Segmentation of T2 weighted Brain MRI images using Convergent Heterogeneous Particle Swarm Optimization

Mohammad Hamed Mozaffari[1], Won-Sook Lee
School of Electrical Engineering and Computer Science, University of Ottawa, 800 King-Edward Avenue, Ottawa, Ontario, Canada, K1N 7S6. Tel: +1-613-562-5800 ext. 2191

## Abstract

This paper proposes a new image thresholding segmentation approach using the heuristic method, Convergent Heterogeneous Particle Swarm Optimization algorithm. The proposed algorithm incorporates a new strategy of searching the problem space by dividing the swarm into subswarms. Each subswarm particles search for better solution separately lead to better exploitation while they cooperate with each other to find the best global position. The consequence of the aforementioned cooperation is better exploration, convergence and it able the algorithm to jump from local optimal solution to the better spots. A practical application of this method is demonstrated for the problem of medical image thresholding segmentation. We considered two classical thresholding techniques of Otsu and Kapur separately as the objective function for the optimization method and applied on a set of brain MR images. Comparative experimental results reveal that the proposed method outperforms another state of the art method from the literature in terms of accuracy, computation time and stable results.

## 1  Introduction

Image processing techniques are used in any fields of science and technology such as astronomy, industry, and electronics. For medicine[1–6], image processing is a vital tool that helps the physicians to interpret images of the patients to diagnose diseases easier and assist the surgeons to perform the operation with more accuracy and effectiveness. Generally, the goal of these techniques is to extract the valuable information in the images that is usually a critical and challenging task. Image segmentation is an absolutely essential preparatory process in almost all image processing approaches in which the image is divided into classes, the background, foreground and region of interest (ROI)[7–9]. As an optimum segmentation, pixels that contribute in each class should be similar in terms of gray- or RGB- level intensities.

Thresholding methods[10,11] are the easiest image segmentation approaches that in their simplest case (bi-level) endeavor to identify one optimum threshold value $t_b$ for the pixels intensity. Thus, the pixels whose intensity values are bigger than $t_b$ are labeled as the first class and the rest labeled as the second class, usually one class is for the background and another for the foreground region of the image[12,13]. Multilevel thresholding is the extended version of the bi-level in which the image pixels are separated into more than two classes corresponding to threshold values.

---

[1] *Corresponding author*
*Email address: mmoza102@uottawa.ca , wslee@uottawa.ca*

According to a prevalent taxonomy, thresholding methods are divided into parametric and nonparametric methods. In the parametric approaches, the gray-level distribution of each class leads to find the thresholds. In order to identify the model of each class, the parameters of the probability density function must be estimated. In the non-parametric techniques, optimizing one criteria such as the entropy, the error rate, and between-class variance results optimal threshold values[12–16]. Time-consuming and computationally expensive of the parametric methods cause that the non-parametric techniques become a desired and common alternative for the thresholding. For better results, criteria in these methods can also be considered as a fitness function for one optimization algorithm which is the main part of this study[17].

Amongst thresholding methods, Otsu's [18–22] and Kapur's [17,23,24] approach are two popular one because of their powerfulness, accuracy, and efficiency especially in the case of bi-level thresholding problem[25]. Although these classical methods perform well in bi-level thresholding problem but for multilevel thresholding problem, because of their exhaustive searching strategy, and greater dimension of the problem space, the computation time grows exponentially with increasing the number of thresholds[26]. Thus, in real-time multilevel thresholding applications, the classical methods are weak and necessity of new techniques is clear. Many studies try to find an alternative way to solve this issue and a common way is to consider multi-level thresholding as an optimization problem.

Heuristic and Meta-heuristic optimization algorithms are known and distinguished for their fast and efficient performance. A combination of heuristic methods and classical thresholding methods not only can solve the problem of computation time, it also can increase the efficiency, accuracy, and robustness of the results[27]. Numerous heuristic techniques successfully have been used for the problem of multilevel thresholding segmentation. Heuristic optimization algorithms are inspired by the laws of physics, mechanics and mimic the developmental process in natural and biological phenomena such as Darwinian evolution in creations (Genetic Algorithm)[15,28–30], crystal and molecular structure shaping in metallurgy (Simulated Annealing)[31], foraging of ants (Ant Colony Optimization)[32], social behaviours of flock of birds or school of fishes in migration (Particle Swarm Optimization)[7,24,33–45], accumulation of water in catchment basins (Inclined Planes system Optimization)[46,47] and numerous more examples.

In order to use aforementioned techniques for thresholding segmentation problem, the common technique is to use the extended version of one classical thresholding method and consider it as a fitness (objective or criteria) function, then heuristic optimization method can find the best threshold values set iteratively by maximizing or minimizing that fitness function [8,28,48–50]. In spite of the fact that heuristic methods are powerful enough in solving high dimensional problems but developing of these methods (better convergence, less sensitive to the parameters, more robust, more accurate and etc.) is still a hot topic for researchers.

In this paper, application of a modified version of the Particle Swarm Optimization (PSO) algorithm with a new strategy of searching, named Convergence Heterogeneous Particle Swarm Optimization (CHPSO)[51] on the problem of multi-level thresholding segmentation of Brain T-2 Magnetic Resonance (MR) images is investigated. Otsu's and Kapur's method is considered as the objective functions and 10 test images are used for the experiments. The experimental results reveal that the proposed method can reach to the best solution within fewer iteration numbers in comparison to the similar approaches without sticking into local solutions.

The rest of the paper is organized as follows: In Section 2, the CHPSO algorithms are introduced with details. In Section 3, the Otsu's and Kapur's methods explained in details and Section 4 shows experimental results and discussion about similar methods as a comparison study. Section 5 concludes the paper.

## 2 Methodology

Heuristic optimization algorithms are designed for solving sophisticated optimization problems (non-linear in most cases) without having knowledge about the problem. Using a reasonable number of constant parameters, computation time and cost lead to an accurate and convergent optimal solution. In almost all heuristic optimization algorithms, there is a trade-off between two methodologies of exploration and exploitation which help the algorithm to search for the best solution efficiently and liberate from capturing in local optima as much as possible. Exploration reinforces algorithm to search the whole range of the problem space by expanding agents (particles) as far as possible in the search space while the exploitation enables the algorithm to find more accurate solutions with better searching locally.

One of the most popular heuristic population-based optimization method is the Particle Swarm Optimization (PSO) which is inspired by the behavior of a flock of birds or a school of fishes during the migration and foraging which their social network helps them to find the best place of food and accommodation. The first version of this algorithm was very powerful and easy to implement, but since 1995 that the PSO algorithm was introduced[33], many improvements have been suggested to update and enhance this method. Convergence Heterogeneous Particle Swarm Optimization (CHPSO) Algorithm[41] is an enhanced version of the PSO which guarantees better optimal solution due to its new strategy of searching and better exploration and exploitation. In the following sections, the original version of the PSO and modified one CHPSO is explained clearly.

### 2.1 Particle Swarm Optimization

In the PSO algorithm, cooperation between particles causes to find the best solution for the search space and each of them can be a candidate for that position. Particles are random selected in the initialization step, then each particle compares its current position with two relative information to update its position and move to a new position, iteratively. The criterion for comparison is the fitness function and the information is the personal best position from the previous iterations and the best position between particles acquired so far.

The position and velocity of particles are calculated respect to this two information in each iteration. Thus, if the position and velocity of $i-th$ particle in the time $t$ and $D$ dimensions are represented by $X_i = [x_i^1, x_i^2, ..., x_i^D]$, and $V_i = [v_i^1, v_i^2, ..., v_i^D]$, respectively, then the position of the particle in the next iteration ($t+1$) is updated according to the following equations.

$$X_i(t+1) = X_i(t) + V_i(t+1) \tag{1}$$

$$V_i(t+1) = \omega V_i(t) + c_1 R(Pbest_i(t) - X_i(t)) + c_2 R(Gbest(t) - X_i(t)) \tag{2}$$

where $c_1$ and $c_2$ are two constants for controlling the exploration and exploitation affects, respectively, $R$ is uniformly distributed random vector from zero to one, $\omega$ is an inertia weight constant which scales the effect of the velocity in the previous iteration, $Pbest_i(t)$ and $Gbest(t)$ are the best personal position and the best position which have been acquired so far (from the time zero to $t$). Note that the equations (1) and (2) are for the global version of the PSO and there is also the local version which is outside the scope of this report.

In a nutshell, for each particle, the corresponding score using the fitness function is calculated then the best personal particle position respect to the previous iteration is stored for using in the next iteration and also the best global particle position respect to the neighbor's particles is selected. This is one of the advantages of the PSO that it needs just one stack of memory and there is no need to calculate complicated formula such as gradient which caused the PSO become a fast and easy to implement the algorithm. The pseudocode of the PSO is tabulated in Algorithm 1.

*Algorithm 1 The PSO algorithm pseudocode*

1. **Initialization.**
    a. **Set** the constant parameters
    b. **Randomize** the vector of positions $X$ and the velocities $V$ of the particles in the search space
    c. **Set** the best personal position to current positions $Pbest = X$
    d. **Set** $Gbest = \arg\{\min f(X)\}$
2. **Termination Check.**
    a. **If** the termination criterion is satisfied stop.
       The $Gbest$ will be the output solution.
    b. **Else** go to Step 3.
3. **For** $i = 1,\ldots,\textit{number of iteration}$ **Do**
    a. Update the velocity and position according to Equation (**2**) and (**1**) and check to be in the range space
    b. Evaluate fitness function of each particle.
    c. Save the best personal and global fitness value and position if better than previous
   **End for**
       Update the $Gbest = \arg\{\min f(Pbest)\}$
4. **Goto** step 2

## 2.2 Convergent Heterogeneous Particle Swarm Optimization

The ideal optimization algorithm with a perfect convergence and accurate solution (not local solutions) in a reasonable computational time and cost is always desirable and attracts many researchers from all fields of science to have a contribution in this area, at least, make an update to enhance previous optimization methods for a particular application. Depends on the search strategy of the optimization algorithm, trapping in local optimum solutions is an inevitable event and less probability of local trapping better results. For the particles of a fast convergent algorithm such as the original PSO, due to its homogeneous searching behavior (diversity of particles is lost) and their weak exploration ability, it is more likely for the algorithm to capture in local optimal spots within the first iterations.

One way to solve this problem is to use a big coefficient for the exploration part of the algorithm at the expense of losing of exploitation effect and losing of accuracy. A new approach to solving this issue is to use multiswarm techniques which are designed to improve the exploration ability of the original heuristic optimization algorithms as well as the exploitation. Cheung et al.[41] in 2014 proposed the CHPSO which uses the multiswarm technique to enhance the weakness of the PSO in jumping from local solutions to the global one.

The main idea of the CHPSO is to divide particles into the four subswarms that each subswarm can search separately the problem space. Each subswarm moves heterogeneously in its local region but also share their information to the other two subswarms to explore new regions and specify the trajectory of the whole swarm in the solution domain. Totally, two subswarms are named basic subswarm which is used for exploitation search and the other two subswarms are called the adaptive subswarm and the exploration subswarm. The adaptive subswarm updates flight path adaptively with using the knowledge that was sent from the two basic subswarms in advanced. Unknown regions in the problem space are discovered by the exploration subswarm which uses the information of the other three subswarms. Because, in CHPSO cooperation between particles is different from the original PSO, new update rules based on positions, velocities and the value of fitness function are defined for CHPSO and are added to the original PSO.

As mentioned before, although each subswarm searches the problem space individually and shares its information with other subswarms (except exploration subswarm), it also contributes in the global exploitation which is the goal of the optimization algorithm. Consequently, all four subswarms search the problem space for the optimum solution heterogeneously while two of them (basic subswarms) supply their velocity and fitness values information for the adaptive subswarm for determining a better trajectory for all particles and finally the velocity information of these three subswarms are shared with the exploration subswarm to enable the algorithm to jump from local optima in order to find the areas of the problem space which never checked.

For better understanding the particles situations, Figure 1 represents a schematic of four subswarms contain five particles that try to find the global best position of the two-dimensional function ( $f(x_1, x_2) = x_1^2 + x_2^2$ ). This cooperation between the particles is used for updating their positions in the next generation. As illustrated in Figure 1, for minimization problem like this, the global position $G_{best}$ is calculated according to the global position in the four subswarms and defined as equation (3).

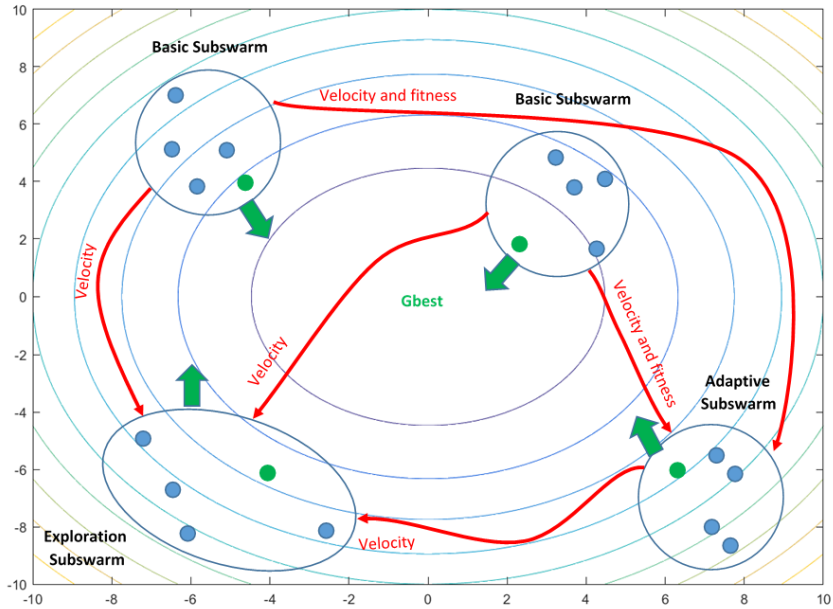

*Figure 1: A schematic of the CHPSO particles and their sharing information model applied on a 2D test function*

$$Gbest = \arg\min\{f(P_{best(sub(s))}) \mid s = 1,2,3,4\} \qquad (3)$$

where, $P_{best(sub(s))}$ is the best particle in subswarms. For updating the position of particles in the adaptive subswarm ( $sub(3)$ ), the velocity and fitness function of the basic subswarms are needed and for updating the exploration subswarm ( $sub(4)$ ) velocity of particles in the basic and adaptive subswarms are needed. Position updating rules for basic subswarms ( $sub(1)$ and $sub(2)$ ) and adaptive subswarm ( $sub(1)$ ) are identical to the original PSO i.e. equations (1). The velocity of the basic subswarms is updated using equation (4).

$$V_{i(sub(1\&2))}(t+1) = \omega V_{i(sub(1\&2))}(t) + c_1 R(P_{i(sub(1\&2))}(t) - X_{i(sub(1\&2))}(t)) + c_2 R(Gbest - X_{i(sub(1\&2))}(t)) \qquad (4)$$

where $c_1$ and $c_2$ are two constants for controlling the exploration and exploitation effects, respectively, $R$ is uniformly distributed random vector from zero to one, $\omega$ is an inertia weight constant which scales the effect of the velocity in the previous iteration, $P_{i(sub(1\&2))}(t)$ and $Gbest$ are the best personal position in the subswarm and the best position which have been gained so far (from the time zero to $t$).

Using the information are acquired from basic subswarms $sub(1)$ and $sub(2)$, particles of the adaptive subswarm $sub(3)$ can adjust the direction of the whole flight. Velocity of the $i-th$ particle in the adaptive subswarm is updated using the following equation:

$$\begin{cases} V_{i(sub(3))}(t+1) = \omega\left(\dfrac{\gamma}{\gamma_1}V_{i(sub(1))}(t+1) + \dfrac{\gamma}{\gamma_2}V_{i(sub(2))}(t+1) + V_{i(sub(3))}(t)\right) + \\ c_1 R(P_{i(sub(3))} - X_{i(sub(3))}(t)) + c_2 R(Gbest - X_{i(sub(3))}(t)) \\ \qquad\qquad \gamma = \gamma_1 + \gamma_2 \end{cases} \qquad (5)$$

where $\gamma_1$ and $\gamma_2$ are the fitness values of the particles in the basic subswarms $sub(1)$ and $sub(2)$, respectively and other parameters are similar to equation (2). The strategy of using $\gamma_1$ and $\gamma_2$ can be explained in this way that the subswarm with better fitness values has a larger effect on the particles in the adaptive subswarm. As discussed, particles of the subswarm $sub(4)$ should attempt to explore new areas as far as possible from other subswarms, thus, the velocity of the $i-th$ particle in the subswarm $sub(4)$ is defined by the difference between the basic subswarms and adaptive one as the equation (6). The $i-th$ particle position in subswarm $sub(4)$ is updated according to its previous information and velocity of the $i-th$ particle in the other subswarm as bellow:

$$V_{i(sub(4))}(t+1) = V_{i(sub(1))}(t+1) + V_{i(sub(2))}(t+1) - V_{i(sub(3))}(t+1) \qquad (6)$$

$$X_{i(sub(4))}(t+1) = \alpha_1 X_{i(sub(4))}(t) + \alpha_2 P_{i(sub(4))} + \alpha_3 Gbest + V_{i(sub(4))}(t+1) \qquad (7)$$

where $\alpha_1$, $\alpha_2$ and $\alpha_3$ are named impact factors which indicate how much the previous information of the particle can contribute in updating process and they must satisfy this equation: $\alpha_1 + \alpha_2 + \alpha_3 = 1$. The bigger impact factor $\alpha_1$ causes the previous information about the $i-th$ particle effect more on updating process than information shared from other subswarms. Generally, diversity and fitness value of the particles in exploration subswarm is considerable because the particles use all the information in the whole swarm. In this paper, the impact factors $\alpha_1$, $\alpha_2$ and $\alpha_3$ are set to $\tfrac{1}{6}$, $\tfrac{1}{3}$ and $\tfrac{1}{2}$, respectively. For better Understanding of the CHPSO algorithm, its pseudocode is illustrated in Algorithm 1.

*Algorithm 2. The CHPSO algorithm pseudocode*

1. **Initialization.**
    a. **Set** the constant parameters
    b. **For** $i = 1,...,4$ **do**
        - **Randomize** the positions of all particles in subswarm $i$ in the search space
        - **Randomize** the velocities of all particles in subswarm $i$ in the search space

> - **Set** the $Gbest_{sub(i)} = \arg\{\min f(x_{isub(i)})\}$
> 
> **End For**
> 
> c. **Set** the swarm global $Gbest = \arg\{\min Gbest_{sub(1,2,3,4)}\}$
> 
> 2. **Termination Check.**
>    a. **If** the termination criterion is satisfied stop.
>       The *Gbest* will be the output solution.
>    b. **Else** go to Step 3.
> 3. **For** $i = 1,\ldots,$ *number of iteration* **Do**
> 
>    **For** $j = 1,\ldots,$ *number of particles* **Do**
>    a. Update the velocity and position of subswarms 1 and 2 according to Equation (4) and (1) and check to be in the range space
>    b. Calculate $\gamma_1$ and $\gamma_2$
>    c. Update the velocity and position of subswarms 3 according to Equation (5) and (1) and check to be in the range space
>    d. Update the velocity and position of subswarms 4 according to Equation (6) and (7) and check to be in the range space
>    e. Evaluate fitness function of each subswarm and particles.
>    f. Save the best subswarm personal and global fitness value and position if better than previous
>    **End for**
>    Update the $Gbest = \arg\{\min Gbest_{sub(1,2,3,4)}\}$
>    **End for**
> 4. **Goto** step 2

In the next chapter, using of the CHPSO for optimizing the Otsu and Kapur criteria is explained.

## 3 Otsu and Kapur Criteria

The aim of image thresholding techniques is to find the optimal threshold values and can be defined and formulated as equations (8) and (9).

$$\text{Bi-level:} \quad \begin{cases} p_{(i,j)} \in S_1 & if \quad 0 \le p_{(i,j)} \prec t_b \\ p_{(i,j)} \in S_2 & if \quad t_b \le p_{(i,j)} \prec L-1 \end{cases} \text{s} \tag{8}$$

$$\text{Multi-level:} \quad \begin{cases} p_{(i,j)} \in S_1 & if \quad 0 \le p_{(i,j)} \prec t_1 \\ p_{(i,j)} \in S_2 & if \quad t_1 \le p_{(i,j)} \prec t_2 \\ p_{(i,j)} \in S_k & if \quad t_k \le p_{(i,j)} \prec t_{k+1} \\ \quad \vdots \\ p_{(i,j)} \in S_m & if \quad t_m \le p_{(i,j)} \prec L-1 \end{cases} \tag{9}$$

where $p_{(i,j)}, i = \{1,2,\ldots,r\}, j = \{1,2,\ldots,c\}$ is the pixel intensity of the image (with size $r \times c$ and with $n = L$ gray levels from zero to $L-1$), $t_k$ is one of the selected different thresholds (in bi-level case $k = b$ with just one threshold value and in the case of multi-level $k = \{1,2,\ldots,L-1\}$ with $m$ number of desired thresholds), $S_k$ is one of the sets in which pixels with intensities between thresholds $t_k$ and $t_{k+1}$ are located.

Otsu[52] proposed a thresholding technique based on probability principles such as zero-, first- and second-order statistics to find the thresholds which can maximize between-class variance. To calculate the probability of each gray-level occurrence in the image, the number of the pixels with that intensities is needed. Histogram diagram is a way to find this numbers as a gray-level distribution. The probability of each gray-level is determined by equation (10).

$$\begin{cases} \Pr_i = \dfrac{h_i}{N_p}, \\ 0 \le i \le L-1, \end{cases} \quad (10)$$

where $h_i$ is the value of histogram (the number of pixels with the specific intensity level $i$), $N_p$ is the number of pixels in the image ($r \times c$). In fact, equation (10) is the normalized version of the histogram by $N_p$ and so $0 \le \Pr_i \prec 1$, $\sum_{i=1}^{N_p} \Pr_i = 1$.

Similar to the Otsu's method, Kapur[23] attempts to find the maximum entropy of the image after segmentation. Normalized Image histogram data again is used as the probability distribution of the pixels intensities. Equations for bi-level and multi-level Otsu's and Kapur's method are tabulated in Table 1 and Table 2, respectively. From the tables, the objective function that is used for the CHPSO is defined by equations (11) and (12) for Otsu's and Kapur's methods respectively.

$$fitness(T) = \max(\sigma^2(T)), \quad (11)$$
$$fitness(T) = \max(H(T)), \quad (12)$$

where vector $T = \{t_1, t_2, \ldots, t_m\}$, $0 \le t_i \le L-1$, $i = 1, 2, \ldots, m$ (for bi-level value $t_b$) is a series of thresholds values.

*Table 1: Otsu's method equations*

| Bi-level | Multilevel |
|---|---|
| Image after thresholding | Image after thresholding |
| Image is divided into two sets $S_1$ and $S_2$ (background and foreground pixels) using a threshold at the level $t_b$, $(m=1)$ | Image after thresholding is divided into $m$ sets $\{S_1, S_2, \ldots, S_m\}$ using thresholds at the levels $\{t_1, t_2, \ldots, t_m\}$ |
| Classes Gray-levels | Classes Gray-levels |
| $S_1 = [0, \ldots, t_b - 1]$<br>$S_2 = [t_b, \ldots, L-1]$ | $S_1 = [0, \ldots, t_1 - 1]$<br>$S_2 = [t_1, \ldots, t_2 - 1]$<br>$\vdots$<br>$S_m = [t_m, \ldots, L-1]$ |
| Probabilities distributions for each class | |
| $\omega(k) = \sum_{i=0}^{k-1} \Pr_i \qquad (13)$ | |
| $\omega_0 = \sum_{i=0}^{t_b-1} \Pr_i = \omega(t_b)$ <br> $\omega_1 = \sum_{i=t_b}^{L-1} \Pr_i = \omega(L) - \omega(t_b)$ <br> $\omega_1 + \omega_2 = 1$    (14) | $\omega_0 = \sum_{i=0}^{t_1-1} \Pr_i = \omega(t_1)$ <br> $\omega_1 = \sum_{i=t_1}^{t_2-1} \Pr_i = \omega(t_2) - \omega(t_1)$ <br> $\vdots$ <br> $\omega_m = \sum_{i=t_m}^{L-1} \Pr_i = \omega(L) - \omega(t_m)$ <br> $\sum_{i=0}^{m} \omega_i = 1$    (15) |
| Mean levels for each class | |
| $\mu(k) = \sum_{i=0}^{k-1} i \Pr_i$    (16) | |
| $\mu_0 = \sum_{i=0}^{t_b-1} \frac{i \Pr_i}{\omega_0} = \frac{\mu(t_b)}{\omega_0}$ <br> $\mu_1 = \sum_{i=t_b}^{L-1} \frac{i \Pr_i}{\omega_1} = \frac{\mu(L) - \mu(t_b)}{\omega_1}$    (17) | $\mu_0 = \sum_{i=0}^{t_1-1} \frac{i \Pr_i}{\omega_0} = \frac{\mu(t_1)}{\omega_0}$ <br> $\mu_1 = \sum_{i=t_1}^{t_2-1} \frac{i \Pr_i}{\omega_1} = \frac{\mu(t_2) - \mu(t_1)}{\omega_1}$ <br> $\vdots$ <br> $\mu_m = \sum_{i=t_m}^{L-1} \frac{i \Pr_i}{\omega_m} = \frac{\mu(L) - \mu(t_m)}{\omega_m}$    (18) |
| Mean levels for the whole image | |
| $\mu_T = \sum_{i=0}^{L-1} i \Pr_i = \mu(L) = \sum_{i=0}^{m} \omega_i \mu_i$    (19) | |
| The Otsu variance between classes | |
| $\sigma^2 = \sigma_0^2 + \sigma_1^2 = \omega_0(\mu_0 - \mu_T)^2 + \omega_1(\mu_1 - \mu_T)^2$    (20) | $\sigma^2 = \sum_{i=0}^{m} \sigma_i = \sum_{i=0}^{m} \omega_i(\mu_i - \mu_T)^2$    (21) |

*Table 2: Kapur's method equations*

| Bi-level | Multilevel |
|---|---|
| Image after thresholding | Image after thresholding |
| Image is divided into two sets $S_1$ and $S_2$ (background and foreground pixels) using a threshold at the level $t_b$, $(m=1)$ | Image after thresholding <br> Image is divided into $m$ sets $\{S_1, S_2, \ldots, S_m\}$ using thresholds at the levels $\{t_1, t_2, \ldots, t_m\}$ |
| Classes Gray-levels | Classes Gray-levels |
| $S_1 = [0, \ldots, t_b - 1]$ <br> $S_2 = [t_b, \ldots, L-1]$ | $S_1 = [0, \ldots, t_1 - 1]$ <br> $S_2 = [t_1, \ldots, t_2 - 1]$ <br> $\vdots$ <br> $S_m = [t_m, \ldots, L-1]$ |
| The Kapur's classes entropies | |

| | | | |
|---|---|---|---|
| $H_0 = \sum_{i=0}^{t_b-1} \frac{\Pr_i}{\omega_0} \ln(\frac{\Pr_i}{\omega_0})$ $H_1 = \sum_{i=t_b}^{L-1} \frac{\Pr_i}{\omega_1} \ln(\frac{\Pr_i}{\omega_1})$ | (22) | $H_0 = \sum_{i=0}^{t_1-1} \frac{\Pr_i}{\omega_0} \ln(\frac{\Pr_i}{\omega_0})$ $H_1 = \sum_{i=t_1}^{t_2-1} \frac{\Pr_i}{\omega_1} \ln(\frac{\Pr_i}{\omega_1})$ $\vdots$ $H_m = \sum_{i=t_m}^{L-1} \frac{\Pr_i}{\omega_m} \ln(\frac{\Pr_i}{\omega_m})$ | (23) |
| The Kapur's overall entropy | | | |
| $H = H_1 + H_2$ | (24) | $H = \sum_{i=0}^{m} H_i$ | (25) |

In this paper, we proposed a combination of the CHPSO and these two criteria to find the best threshold values of the test image. In the next chapter, the application of the proposed method for medical image threshold segmentation is investigated and compared to the similar method.

# 4 Experimental results and Discussion

In order to evaluate the performance of the CHPSO, it has been applied on 10 benchmark images and the results compared with a novel heuristic algorithm named Amended Bacterial Foraging (ABF) algorithm[53]. Images are T2-weighted Brain-hemispheric MR and downloaded from the whole brain atlas database[54]. Transaxial slices from 22 to 112 with 10 slice gap interval are selected from the database. All the images have the same size ($256 \times 256$ pixels) and they are in PNG format. A representation of the test images is depicted in Figure 2.

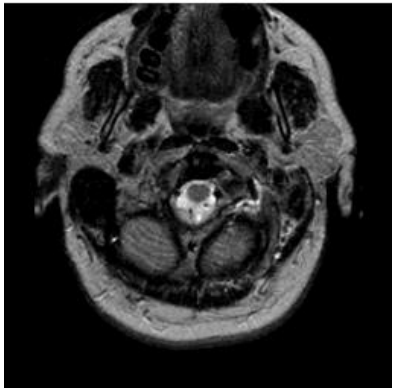 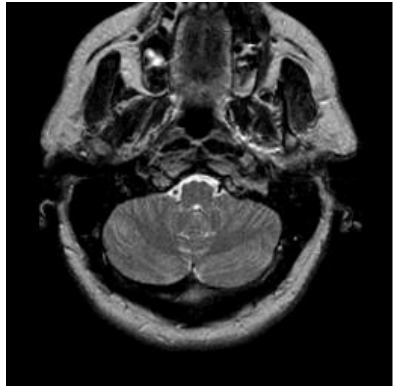 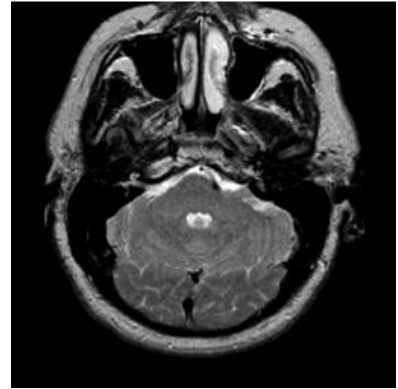

(a)           (b)           (c)

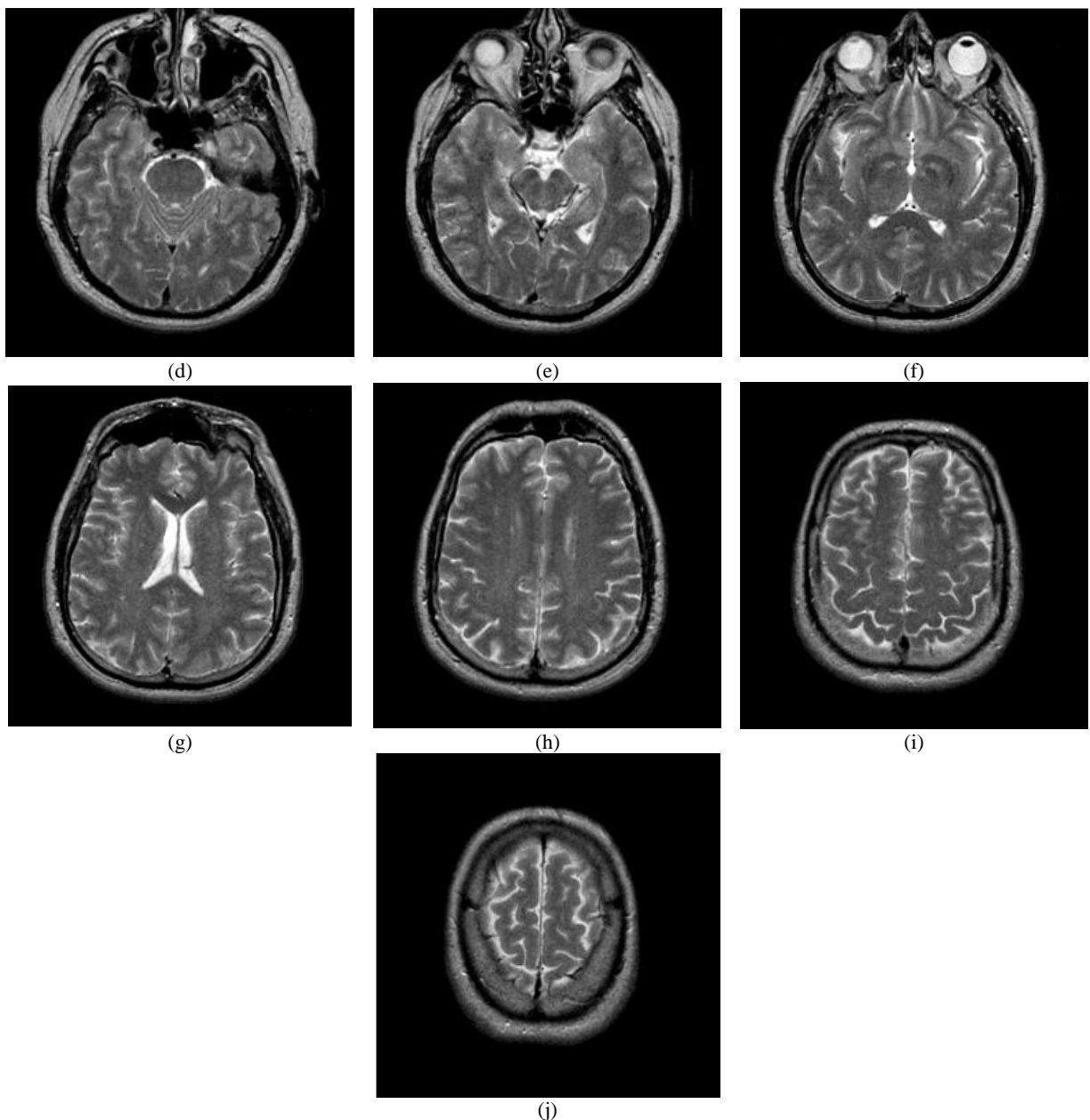

*Figure 2. Test images: A to J are Slices 22, 32, 43, 52, 63, 72, 82, 92, 102, 112, respectively.*

*Since random numbers are used in the CHPSO algorithm like other heuristic methods in each run of the algorithm the outcome most likely to different from another run. Thus, it is valuable to employ an appropriate statistical metrics to measure the efficiency of the method. The results have been reported after executing the algorithm for 50 times for each image. It is common in the literature that the number of threshold points is selected from numbers 2, 3, 4, and 5. The number of iterations is set as stopping criteria for the CHPSO and there are not any limit for the best fitness values remains with no change. Therefore, algorithm run for the whole of the time interval and the parameters of the method are tabulated in*

Table 3. We try to make a similar situation to the ABF method for better comparison.

*Table 3. Parameters settings for CHPSOMT*

| Number of iteration | Number of particles | Acceleration constants | Inertia weight |
|---|---|---|---|
| 100 | 20 | 1.49 for both $c_1$ and $c_2$ | Linearly is changed from 0.4 to 0.9 |

Calculating the Standard Deviation (STD) from equation (26) is a desirable way to show the dispersion of the data after 50 times repetition and minimum STD for algorithms the better stability.

$$STD = \sqrt{\left(\sum_{i=1}^{N} \frac{Bestfit_i - Meanfit}{N}\right)} \quad (26)$$

where $N$ is the number of algorithm run $(N=50)$, $bestfit_i$ is the best fitness acquired from the $i-th$ algorithm execution and $Meanfit$ is the average result of all best finesses. Peak Signal to Noise Ratio (PSNR) is another way of assessing the proposed method in respect to quality and noise effects on the image. It compares the threshold delineated image with original one as a reference to find how much of original data are conveyed to the segmented image. The bigger PSNR the better signal quality and for defining the PSNR a common way is first to determine the mean squared error (MSE) by using both image data from the equation (27), then PSNR in decibels unit (dB) can be calculated by equation (28)[17].

$$MSE = \frac{1}{r \times c} \sum_{i=0}^{r-1} \sum_{j=0}^{c-1} [I(i,j) - I_s(i,j)]^2 \quad (27)$$

$$PSNR = 20\log_{10}\left(\frac{255}{\sqrt{MSE}}\right) \quad (28)$$

where $r$ and $c$ are the number of row and columns of gray-level original and segmented images ($I$ and $I_s$), respectively. In order to qualitatively assessment of the method, the results can be evaluated by the popular uniformity measure as equation (29)[53].

$$u = 1 - 2 \times m \times \frac{\sum_{j=0}^{m} \sum_{i \in R_j} (f_i - \mu_j)^2}{N \times (f_{max} - f_{min})^2} \quad (29)$$

where, $m$ is the number of thresholds, $R_j$ is the $j-th$ region of the image that is segmented, $f_i$ is the gray level of the pixel $i$, $\mu_j$ is the mean of the gray levels inside the region $R_j$, $N$ is the number of total pixels in the image, $f_{min}$ and $f_{max}$ are the minimum and maximum gray level of pixels within the image respectively. The value of the uniformity measure is between 0 and 1 [29]. Uniformity with higher value means the better thresholding. Also for having a comparison with ABF results, misclassification error (in present) which is the difference between the best thresholding with $u=1$ and the value calculated from equation (29) is considered as a criterion.
The CHPSO is applied over the complete set of benchmark images first considering the Otsu's method (equation (12) as the fitness function) and then using the Kapur's method as the fitness function (equation (11)) to find the optimum thresholds. Note that, because of the CHPSO optimization algorithm has the type of minimization, fitness functions values are multiplied by minus to change the problem space from maximization to minimization.

A visualization of the results after applying the proposed method on the images are illustrated in Figure 3 and Figure 4 using Otsu's and Kapur's criteria, respectively. As we can see in the figures, the white matter, gray matter, and cerebrospinal fluid are delineated perfectly with more details. In the case of the bigger number of thresholds, the quality of the image is significantly considerable and reveals that the proposed method can save similar patterns in the original image and convey them to the segmented one. In medical applications, the details in the images mean better diagnosis and better treatment and the images proof that our method can be used for this purposes.

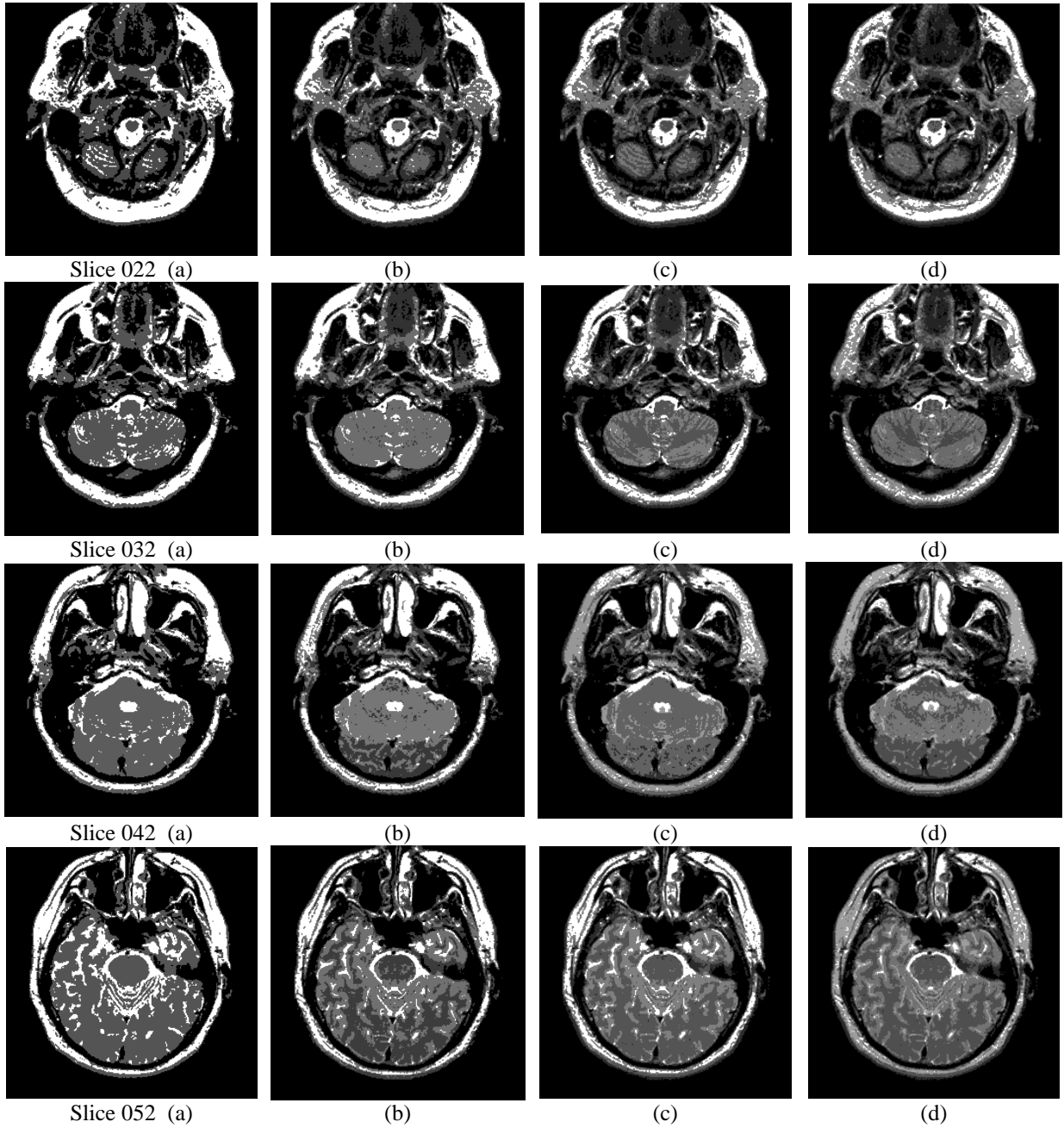

Slice 022  (a)     (b)     (c)     (d)

Slice 032  (a)     (b)     (c)     (d)

Slice 042  (a)     (b)     (c)     (d)

Slice 052  (a)     (b)     (c)     (d)

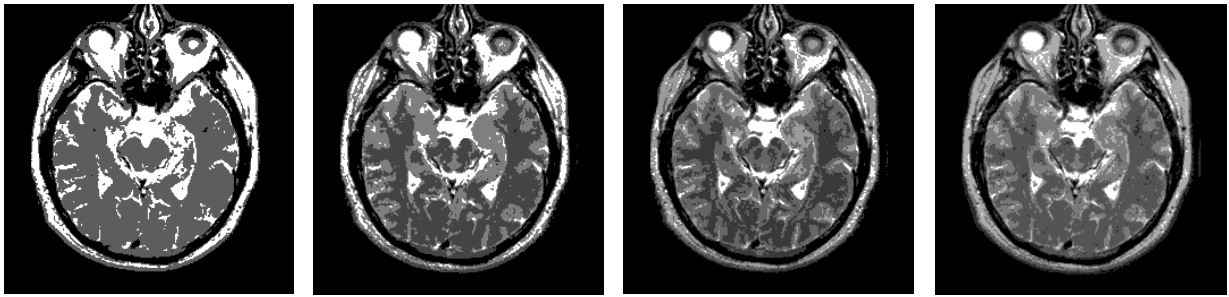
Slice 062 (a)     (b)     (c)     (d)

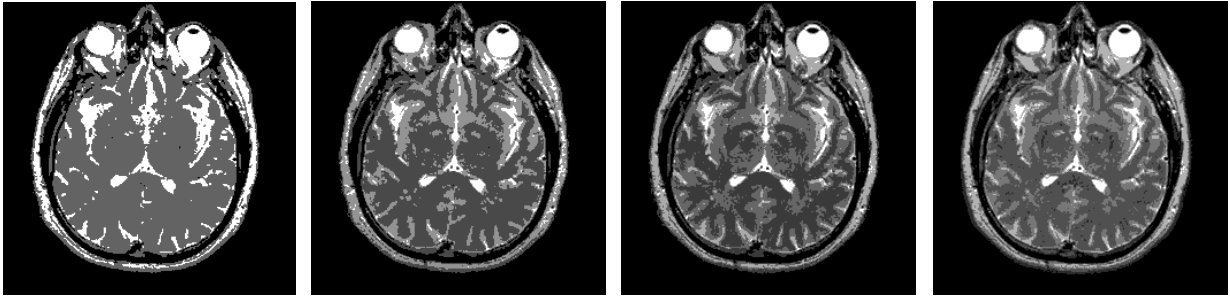
Slice 072 (a)     (b)     (c)     (d)

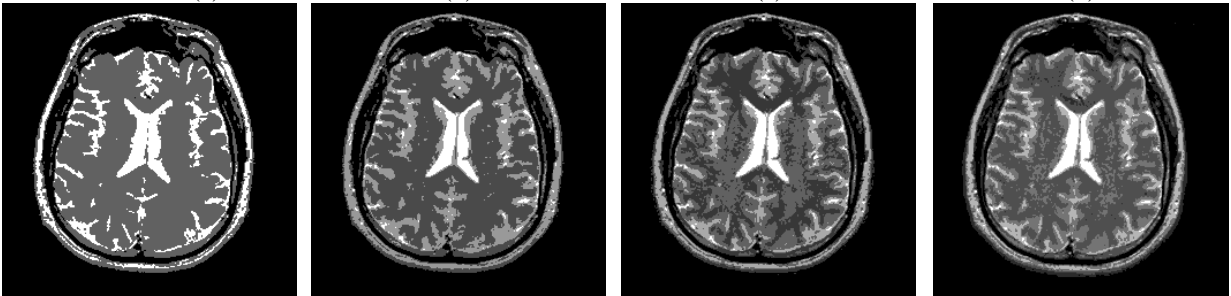
Slice 082 (a)     (b)     (c)     (d)

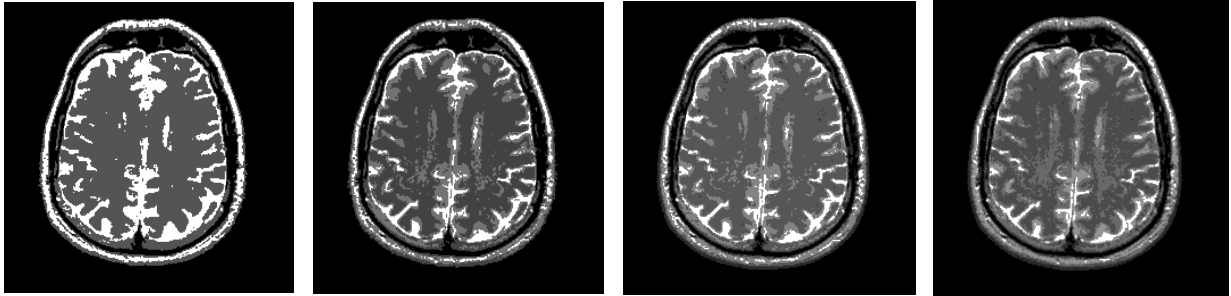
Slice 092 (a)     (b)     (c)     (d)

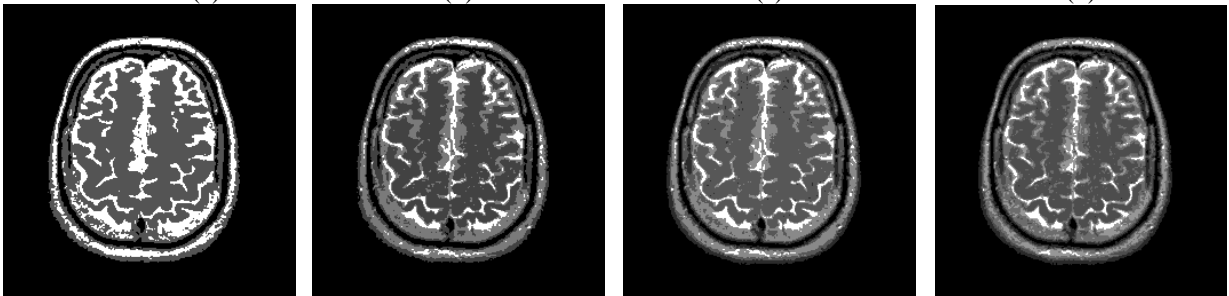
Slice 102 (a)     (b)     (c)     (d)

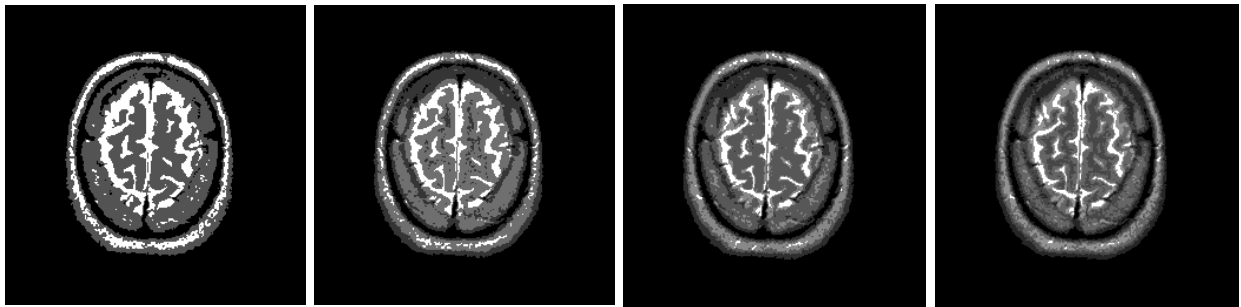
Slice 112  (a)　　　　　　(b)　　　　　　(c)　　　　　　(d)
*Figure 3. CHPSO results using Otsu's method – (a to d) are images after thresholding with m = 2, 3, 4 and 5, respectively*

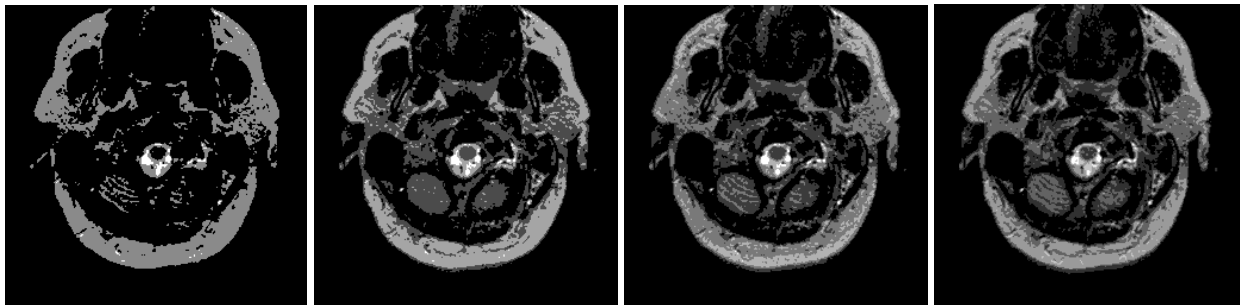
Slice 022  (a)　　　　　　(b)　　　　　　(c)　　　　　　(d)

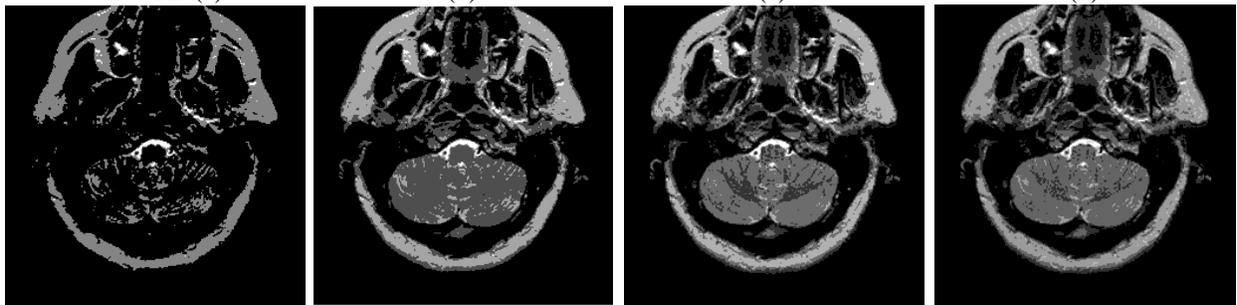
Slice 032  (a)　　　　　　(b)　　　　　　(c)　　　　　　(d)

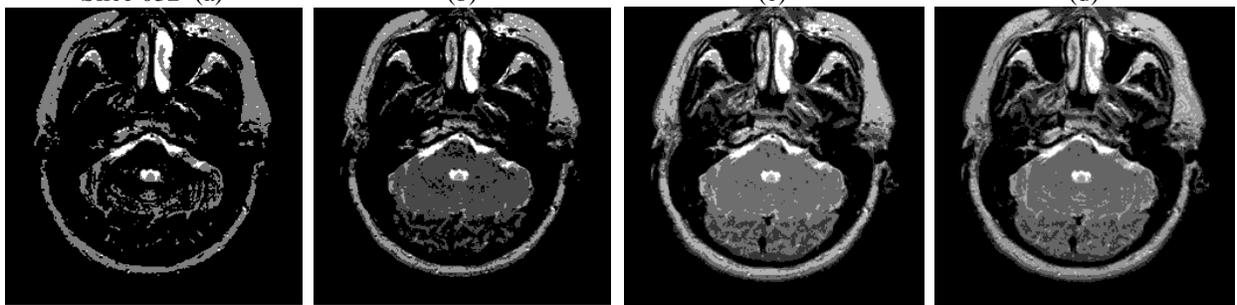
Slice 042  (a)　　　　　　(b)　　　　　　(c)　　　　　　(d)

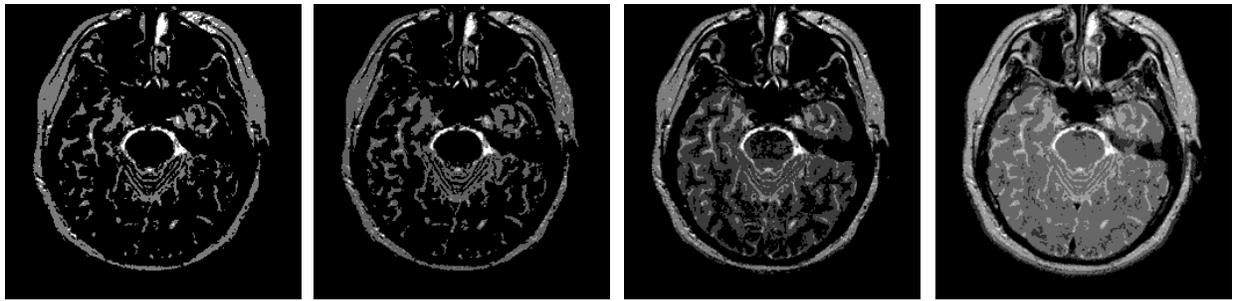

Slice 052  (a)　　　　(b)　　　　(c)　　　　(d)

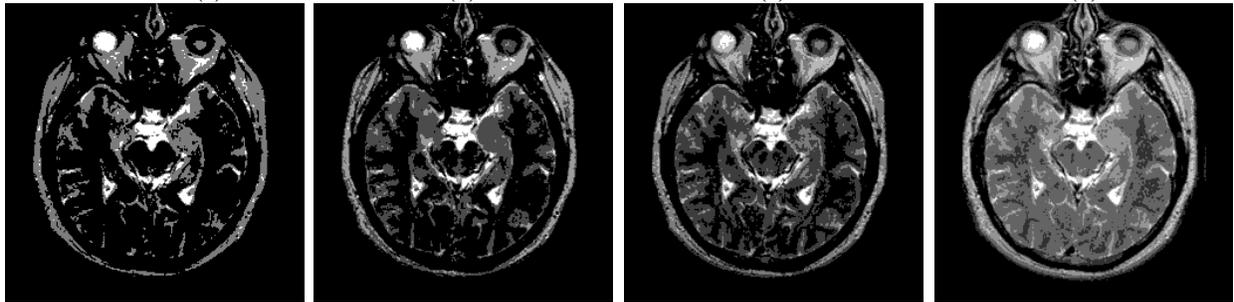

Slice 062  (a)　　　　(b)　　　　(c)　　　　(d)

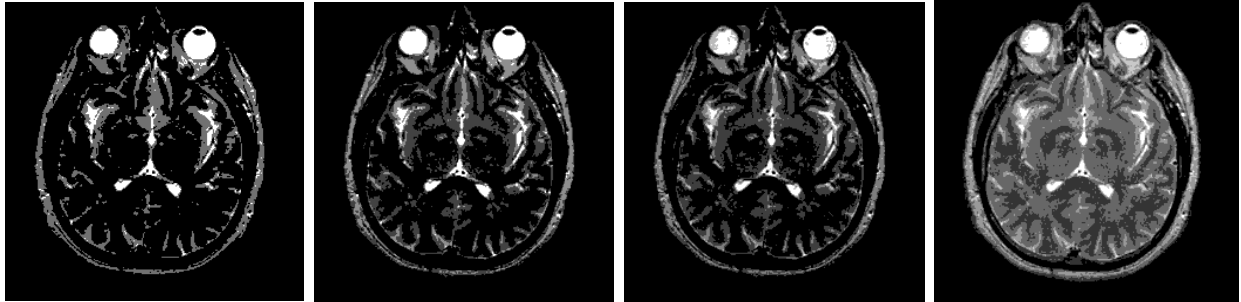

Slice 072  (a)　　　　(b)　　　　(c)　　　　(d)

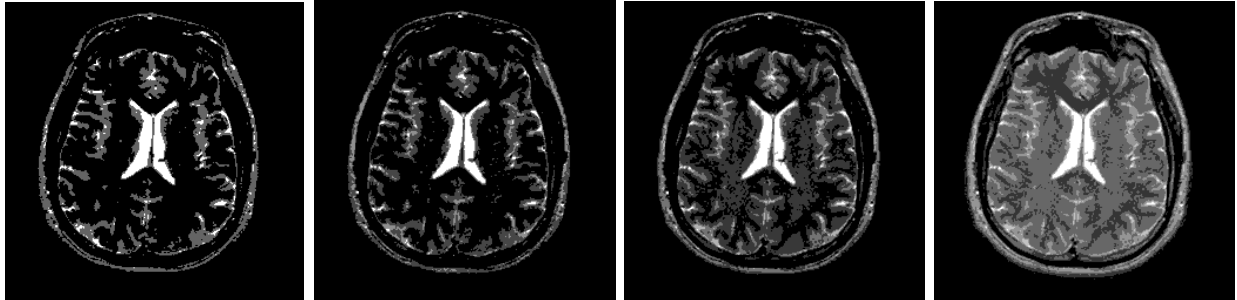

Slice 082  (a)　　　　(b)　　　　(c)　　　　(d)

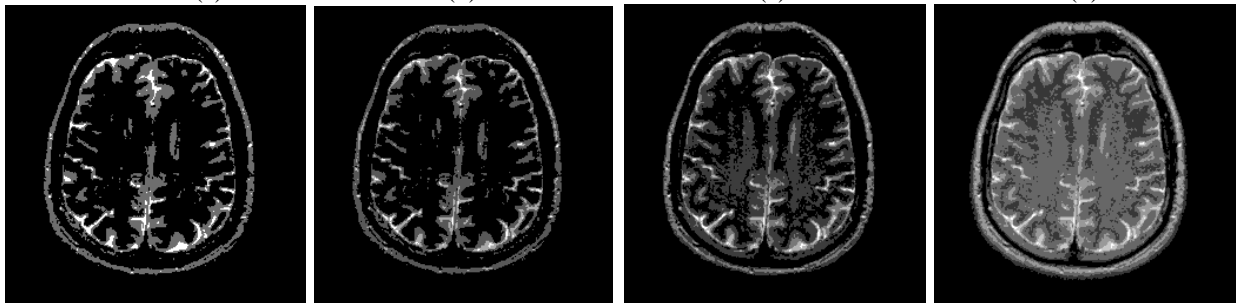

Slice 092  (a)　　　　(b)　　　　(c)　　　　(d)

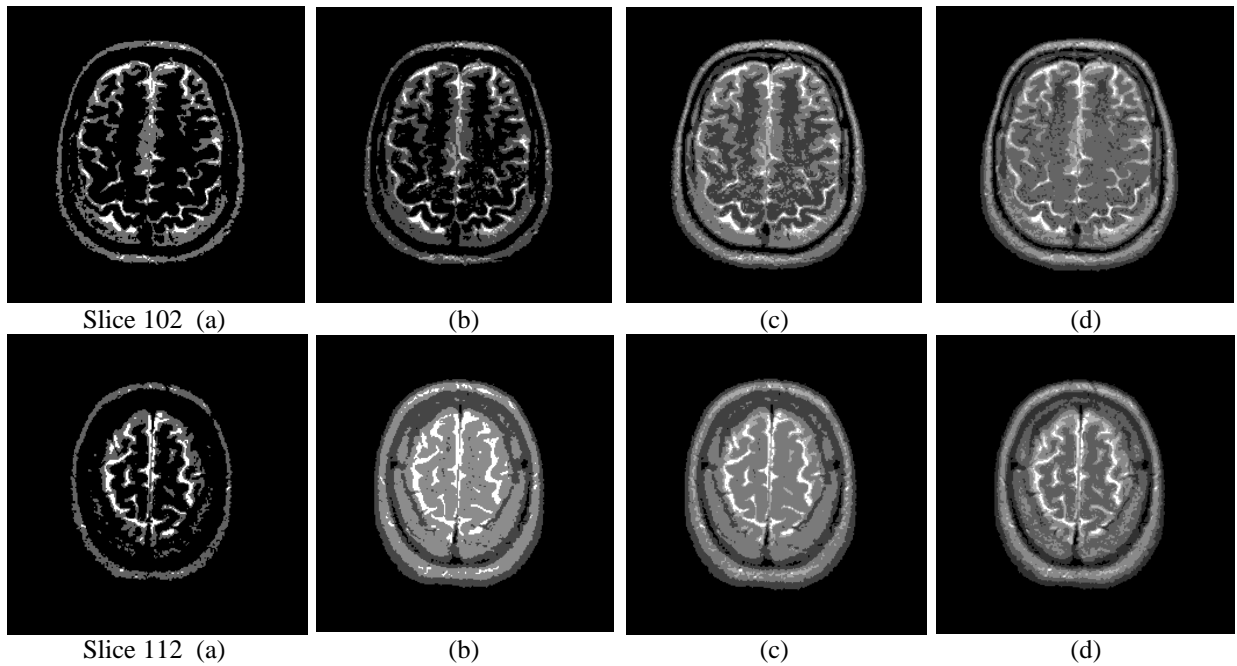

Slice 102 (a)     (b)     (c)     (d)

Slice 112 (a)     (b)     (c)     (d)

*Figure 4. CHPSO results using Kapur's method – (a to d) are images after thresholding with m = 2, 3, 4 and 5, respectively*

The performance of the CHPSO for image thresholding was astonishing and clearly shows the powerfulness of this method, but this results should be compared with other similar techniques to figure out the advantages and disadvantages of the CHPSO. For this reason, the results of the ABF algorithm are tabulated beside the CHPSO in the following tables.

From the Table 4, we can assert that two methods have a similar amount of fitness function, especially for Kapur's method. In the case of Otsu, the results of the two proposed method are different but they could converge well. The difference between this results is that no pre- or post-processing have been done on the images for the case of CHPSO. For instance, we can see threshold value of 0 in the slice 112 that is because in the original image there is a line with zero intensity.

*Table 4. Comparison study of threshold and fitness function values obtained by CHPSO and ABF using Otsu and Kapur*

| image | m | CHPSO | | | | ABF | | | |
|---|---|---|---|---|---|---|---|---|---|
| | | Otsu | | Kapur | | Otsu | | Kapur | |
| | | Optimum Thresholds values | Fitness function | Optimum Thresholds values | Fitness function | Optimum Thresholds values | Fitness function | Optimum Thresholds values | Fitness function |
| **Slice22** | 2 | 35,99 | 2273.7 | 94,182 | 9.2155 | 93,176 | 1808.8478 | 95,184 | 9.2155 |
| | 3 | 22,66,117 | 2370.3 | 56,114,184 | 11.733 | 68,118,188 | 2165.1669 | 68,115,186 | 11.7123 |
| | 4 | 16,51,88,128 | 2409.3 | 47,96,146,192 | 13.946 | 45,109,160,184 | 2288.3476 | 43,89,136,186 | 13.9481 |
| | 5 | 14,42,73,105,140 | 2429.2 | 41,83,124,170,205 | 16.105 | 47,90,128,174,191 | 2321.4811 | 44,85,127,173,207 | 16.2228 |
| **Slice32** | 2 | 43,112 | 2607.1 | 108,184 | 9.2645 | 107,178 | 1809.3391 | 110,185 | 9.2644 |
| | 3 | 23,69,122 | 2712.4 | 52,114,184 | 11.683 | 68,125,168 | 2557.1341 | 82,130,190 | 11.6166 |
| | 4 | 18,58,96,135 | 2751.8 | 38,84,132,188 | 13.936 | 51,94,129,188 | 2672.3691 | 44,92,136,188 | 13.9222 |
| | 5 | 17,51,84,116,150 | 2773.8 | 33,77,121,168,203 | 16.043 | 45,79,115,152,194 | 2714.2533 | 42,92,134,184,216 | 16.0040 |
| **Slice42** | 2 | 45,121 | 3044.4 | 112,182 | 9.2585 | 112,182 | 2118.4914 | 114,184 | 9.2585 |
| | 3 | 30,81,135 | 3143.9 | 82,130,186 | 11.578 | 78,127,178 | 2891.0697 | 83,136,187 | 11.5743 |
| | 4 | 23,67,113,161 | 3199.7 | 28,75,126,186 | 13.865 | 57,112,138,177 | 3120.0412 | 48,102,144,190 | 13.8155 |
| | 5 | 18,55,91,128,174 | 3234.6 | 26,71,116,162,201 | 16.026 | 53,90,126,162,202 | 3166.4774 | 40,93,137,183,217 | 15.9644 |
| **Slice52** | 2 | 45,115 | 2858.8 | 116,184 | 9.2447 | 118,188 | 1569.4257 | 117,186 | 9.2447 |
| | 3 | 36,88,132 | 2946.9 | 108,164,202 | 11.58 | 98,130,176 | 2099.4181 | 109,163,203 | 11.5552 |
| | 4 | 19,58,97,137 | 2997.4 | 83,124,167,206 | 13.734 | 84,109,140,180 | 2488.3868 | 92,131,173,210 | 13.7502 |
| | 5 | 18,57,92,125,164 | 3023.6 | 25,69,117,165,203 | 15.874 | 59,99,129,162,206 | 2934.7700 | 48,89,131,175,209 | 15.7152 |
| **Slice62** | 2 | 46,123 | 3369.4 | 120,186 | 9.3367 | 118,183 | 2158.5697 | 119,186 | 9.3314 |
| | 3 | 39,96,147 | 3484.6 | 100,146,194 | 11.674 | 94,132,186 | 2781.2290 | 99,150,195 | 11.6688 |
| | 4 | 35,84,122,169 | 3537.5 | 86,126,169,207 | 13.77 | 78,118,151,198 | 3217.2777 | 93,134,176,212 | 13.7812 |
| | 5 | 19,58,96,132,179 | 3577.9 | 28,72,117,161,201 | 15.866 | 75,102,130,162,201 | 3312.1446 | 82,114,147,181,216 | 15.6953 |
| **Slice72** | 2 | 45,126 | 3206.2 | 116,178 | 9.4205 | 111,173 | 2082.9210 | 117,179 | 9.4205 |

| | 3 | 40,103,164 | 3342.1 | 98,140,186 | 11.694 | 103,145,196 | 2270.0114 | 99,145,196 | 11.6758 |
| | 4 | 36,87,124,182 | 3405.5 | 95,134,174,213 | 13.846 | 99,126,159,201 | 2452.8171 | 99,140,179,214 | 13.9463 |
| | 5 | 21,60,94,130,185 | 3441.5 | 47,88,129,169,208 | 15.797 | 75,100,130,168,209 | 3137.3314 | 84,116,148,179,215 | 15.8427 |
| **Slice82** | 2 | 45,125 | 2938.1 | 110,168 | 9.191 | 115,167 | 1653.4012 | 111,170 | 9.1910 |
| | 3 | 42,105,171 | 3061.3 | 102,144,188 | 11.427 | 108,145,200 | 1826.7544 | 103,146,194 | 11.4140 |
| | 4 | 35,83,119,181 | 3117.1 | 84,122,161,203 | 13.508 | 98,126,145,198 | 2111.0125 | 95,131,170,210 | 13.6188 |
| | 5 | 19,58,92,127,187 | 3151.6 | 33,79,123,164,202 | 15.564 | 87,107,131,162,212 | 2505.4256 | 91,123,158,189,220 | 15.4033 |
| **Slice92** | 2 | 43,113 | 2654 | 108,172 | 8.7906 | 100,182 | 1612.4944 | 109,174 | 8.7906 |
| | 3 | 39,95,136 | 2716.1 | 104,156,204 | 11.164 | 97,141,193 | 1671.3594 | 105,159,207 | 11.1560 |
| | 4 | 19,59,99,140 | 2750.7 | 85,126,165,206 | 13.276 | 90,120,156,191 | 1971.2016 | 97,135,174,212 | 13.2974 |
| | 5 | 18,56,89,116,155 | 2775.5 | 29,78,120,165,206 | 15.376 | 88,110,137,166,208 | 1993.7386 | 92,123,155,186,216 | 15.2733 |
| **Slice102** | 2 | 43,112 | 2571.6 | 106,172 | 8.5283 | 100,158 | 1732.1681 | 108,174 | 8.5283 |
| | 3 | 39,95,142 | 2643.6 | 92,140,188 | 10.928 | 93,139,184 | 1851.2240 | 95,143,190 | 10.9375 |
| | 4 | 20,60,101,146 | 2682.7 | 31,85,135,182 | 13.067 | 91,126,147,206 | 2000.5093 | 89,132,170,204 | 13.0186 |
| | 5 | 19,57,90,119,158 | 2703.9 | 16,71,111,151,193 | 15.31 | 84,110,139,179,206 | 2127.9426 | 85,122,154,186,217 | 14.8850 |
| **Slice112** | 2 | 41,110 | 2016.6 | 104,162 | 8.1476 | 93,170 | 1843.8040 | 105,164 | 8.1476 |
| | 3 | 26,74,127 | 2090.2 | 1,70,141 | 10.601 | 78,124,181 | 1907.0305 | 74,130,176 | 10.4500 |
| | 4 | 22,65,102,147 | 2126.7 | 0,65,122,170 | 13.058 | 72,104,144,182 | 1973.3334 | 47,96,141,184 | 12.7611 |
| | 5 | 19,55,84,112,154 | 2141.1 | 0,49,95,138,182 | 15.281 | 60,95,130,168,196 | 2039.0589 | 50,99,140,175,208 | 14.6592 |

Although, the results for the Otsu's methods is different but we can compare them in terms of statistical analysis.

In terms of the Standard deviation, the Peak Signal to Noise Ratio, we can understand from Table 5 that the CHPSO is highly robustness and in almost all cases gained a better STD rather than ABF and consequently more trustable respect to changes of output in each execution of the algorithm. The difference of PSNR for all images about 2 times bigger, reveal another property of the proposed method. The performance of the CHPSO in skipping from the local minima to better solutions lead to outstanding results in compare to the ABF and proof the power of multiswarm optimization algorithms for better exploration of the search space.

*Table 5. Comparison study of the CHPSO and ABF in terms of STD and PSNR*

| image | m | CHPSO | | | | ABF | | | |
| --- | --- | --- | --- | --- | --- | --- | --- | --- | --- |
| | | Otsu | | Kapur | | Otsu | | Kapur | |
| | | STD | PSNR | STD | PSNR | STD | PSNR | STD | PSNR |
| **Slice22** | **2** | **0** | **24.5006** | **1.0766e-14** | **24.269** | 0.0021 | 10.0804 | 1.1721e-4 | 10.3797 |
| | **3** | **2.2968e-12** | **24.6694** | **5.3832e-15** | **24.416** | 0.7238 | 13.4236 | 0.0009 | 13.4504 |
| | **4** | **1.3781e-12** | **24.8633** | 0.011992 | **24.573** | 0.8222 | 15.5969 | **0.0021** | 16.2311 |
| | 5 | 3.1395 | **25.6535** | 0.049308 | **24.654** | **1.0210** | 16.8073 | **0.0165** | 17.7098 |
| **Slice32** | **2** | **2.7562e-12** | **24.4469** | **7.1776e-15** | **24.315** | 0.3119 | 9.1680 | 1.0022e-4 | 9.2958 |
| | **3** | **1.8375e-12** | **24.5709** | **1.0766e-14** | **24.444** | 0.4844 | 12.5778 | 0.0018 | 13.2597 |
| | **4** | **3.6749e-12** | **24.7943** | 0.023538 | **24.567** | 0.5001 | 15.5448 | **0.0111** | 16.9498 |
| | 5 | 0.18787 | 24.9692 | 0.013657 | **24.674** | 0.5313 | 17.2658 | 0.0329 | 17.3335 |
| **Slice42** | **2** | **1.3781e-12** | **24.3693** | **8.972e-15** | **24.339** | 0.2813 | 9.0416 | 1.5000e-4 | 9.0892 |
| | **3** | **0** | **24.5606** | **8.972e-15** | **24.443** | 0.4063 | 12.4376 | 0.0175 | 12.9165 |
| | **4** | 0.016243 | **24.7001** | **0.00032519** | **24.585** | 0.7813 | 15.0468 | 0.0527 | 16.3489 |
| | 5 | 0.0082017 | **24.8189** | 0.0073274 | **24.68** | 1.2969 | 15.9160 | 0.0729 | 17.7189 |
| **Slice52** | **2** | **9.1873e-13** | **24.4490** | **5.3832e-15** | **24.359** | 0.1278 | 9.1512 | 1.3018e-4 | 9.2423 |
| | **3** | **9.1873e-13** | **24.6998** | **8.972e-15** | **24.479** | 0.5000 | 9.9752 | 0.0015 | 10.3772 |
| | **4** | **0.0042822** | **24.7770** | 0.048056 | **24.661** | 0.9375 | 11.4083 | 0.068 | 11.9692 |
| | 5 | 0.1792 | **24.9605** | 0.046858 | **24.698** | 1.7656 | 15.4337 | 0.0996 | 16.8916 |
| **Slice62** | **2** | **4.5936e-12** | **24.3949** | **3.5888e-15** | **24.354** | 0.1474 | 9.1676 | 1.1012e-4 | 9.2923 |
| | **3** | **2.2968e-12** | **24.6499** | **5.3832e-15** | **24.574** | 1.0001 | 10.7943 | 0.0047 | 11.0396 |
| | **4** | **3.6749e-12** | **24.8458** | 0.036804 | **24.718** | 1.2031 | 11.1523 | **0.0169** | 11.9759 |
| | 5 | 0.66916 | **24.8808** | 0.080143 | **24.77** | 1.8906 | 13.8409 | 0.0867 | 14.8511 |
| **Slice72** | **2** | **2.7562e-12** | **24.3542** | **8.972e-15** | **24.358** | 0.2594 | 8.9376 | 5.5610e-5 | 9.4718 |
| | **3** | **2.2968e-12** | **24.5795** | **8.972e-15** | **24.597** | 0.5156 | 10.7279 | 0.0097 | 11.0624 |
| | **4** | **1.3781e-12** | **24.8080** | **0.00020507** | **24.656** | 0.6875 | 11.1622 | 0.0378 | 11.7246 |
| | 5 | 0.017521 | **24.8603** | 0.053892 | **24.794** | 1.0625 | 11.8284 | 0.0642 | 12.6370 |
| **Slice82** | **2** | **2.7562e-12** | **24.3593** | **5.3832e-15** | **24.427** | 0.2731 | 9.4512 | 2.7307e-4 | 9.7528 |
| | **3** | **4.5936e-13** | **24.5585** | **3.5888e-15** | **24.59** | 1.1406 | 10.0912 | 0.0049 | 10.5192 |

| | 4 | 9.1873e-13 | 24.7953 | 0.027152 | 24.741 | 1.5781 | 11.0186 | 0.0318 | 11.2897 |
| --- | --- | --- | --- | --- | --- | --- | --- | --- | --- |
| | 5 | 0.014472 | 24.8838 | 0.11362 | 24.716 | 1.8594 | 12.2378 | **0.0508** | 12.7024 |
| | 2 | 3.6749e-12 | 24.3605 | 8.972e-15 | 24.365 | 0.5419 | 9.2776 | 1.6631e-4 | 9.7164 |
| | 3 | 1.8375e-12 | 24.7015 | 0 | 24.46 | 1.3438 | 9.8622 | 0.0117 | 10.2338 |
| **Slice92** | 4 | 1.3781e-12 | 24.6962 | 0.05869 | 24.699 | 2.0000 | 10.5923 | **0.0248** | 10.9162 |
| | 5 | 0.017778 | 24.9622 | 0.13649 | 24.68 | 2.4375 | 10.8404 | **0.0438** | 11.4210 |
| | 2 | 3.2155e-12 | 24.3440 | 8.972e-15 | 24.327 | 0.2828 | 9.3085 | 5.9248e-5 | 9.3821 |
| | 3 | 2.7562e-12 | 24.5524 | 0 | 24.512 | 1.0056 | 10.5222 | 0.0031 | 10.6334 |
| **Slice102** | 4 | 1.3781e-12 | 24.5521 | 0.091335 | 24.672 | 1.1406 | 11.1791 | **0.0045** | 11.2439 |
| | 5 | 0.78066 | 24.7973 | 0.21657 | 24.914 | 1.2656 | 11.8127 | **0.0099** | 12.6854 |
| | 2 | 1.1484e-12 | 24.3324 | 5.3832e-15 | 24.296 | 0.2920 | 9.0078 | 0.0022 | 9.0723 |
| | 3 | 1.3781e-12 | 24.3872 | 0.013084 | 29.42 | 0.8281 | 12.5501 | **0.0057** | 12.8712 |
| **Slice112** | 4 | 1.3781e-12 | 24.5038 | 0.0019779 | 29.672 | 1.5469 | 12.8380 | 0.0098 | 15.9141 |
| | 5 | 0.020992 | 24.6291 | 0.0090181 | 30.25 | 1.8156 | 14.9048 | 0.0367 | 16.5306 |

From the **Error! Not a valid bookmark self-reference.**, we can see that the CHPSO is also significantly faster than ABF algorithm. In this report, the algorithm is implemented using MATLAB software on the PC with a dual core of 3GHz with 2 GB memory. Faster results mean the ability of the proposed method for using in the real-time application that is common in medicine.

*In term of misclassification error in percentage, CHPSO could achieve better results in most cases especially when the number of thresholds is five. For the sake of representation the results of only three and five thresholds values are illustrated in*

Table 7.

In general, the experimental results reveals that the CHPSO approach using both Otsu and Kapur's criteria is a robust, fast and accurate for the problem of medical image thresholding segmentation in comparison to the similar methods. The CHPSO for getting perfect results owe to the multi-swarm searching strategy and the behavior of the particles that able them to search more new regions in the problem space and jump from local minima to a global one and discover a bigger area. For the speed of the algorithm, it was predictable, because of the simple equations of the original PSO that have not any complex calculations.

*Table 6. Comparison study of CHPSO and ABF in term of computation time (Ct) in second*

| image | m | CHPSO | | ABF | |
| --- | --- | --- | --- | --- | --- |
| | | Ct Otsu | Ct Kapur | Ct Otsu | Ct Kapur |
| **Slice22** | 2 | **0.1358** | **0.28989** | 3.0993 | 6.5245 |
| | 3 | **0.1353** | **0.29003** | 3.2681 | 6.9201 |
| | 4 | **0.1334** | **0.27506** | 3.7904 | 7.4347 |
| | 5 | **0.1313** | **0.26038** | 3.9115 | 8.1779 |
| **Slice32** | 2 | **0.1375** | **0.28806** | 2.7974 | 6.7466 |
| | 3 | **0.1347** | **0.28466** | 3.2821 | 7.2019 |
| | 4 | **0.1352** | **0.27242** | 3.9708 | 7.5040 |
| | 5 | **0.1313** | **0.2554** | 4.6637 | 8.1982 |
| **Slice42** | 2 | **0.1347** | **0.28964** | 2.7200 | 6.7982 |
| | 3 | **0.1347** | **0.27923** | 3.2588 | 7.1599 |
| | 4 | **0.1347** | **0.27337** | 3.8201 | 7.8631 |
| | 5 | **0.1333** | **0.255** | 4.5712 | 8.1029 |
| **Slice52** | 2 | **0.1362** | **0.28743** | 2.7831 | 6.7001 |
| | 3 | **0.1352** | **0.28174** | 3.3111 | 7.0938 |
| | 4 | **0.1346** | **0.27972** | 3.6868 | 7.2184 |
| | 5 | **0.1313** | **0.25377** | 4.4951 | 8.2670 |
| **Slice62** | 2 | **0.1353** | **0.30375** | 2.9800 | 6.5281 |
| | 3 | **0.1351** | **0.27951** | 3.2259 | 6.8914 |
| | 4 | **0.1341** | **0.28061** | 3.2801 | 7.1803 |
| | 5 | **0.1301** | **0.25346** | 4.8114 | 8.1060 |
| **Slice72** | 2 | **0.1344** | **0.29204** | 2.7206 | 6.6929 |
| | 3 | **0.1361** | **0.27807** | 3.0875 | 7.3030 |

| | 4 | 0.1373 | 0.26924 | 3.8888 | 7.4115 |
| | 5 | 0.1324 | 0.25184 | 4.6753 | 8.3934 |
| **Slice82** | 2 | 0.1354 | 0.28678 | 2.8140 | 6.6602 |
| | 3 | 0.1347 | 0.28047 | 3.1250 | 7.4265 |
| | 4 | 0.1347 | 0.26758 | 3.6752 | 7.5396 |
| | 5 | 0.1322 | 0.24793 | 4.2175 | 7.8576 |
| **Slice92** | 2 | 0.1352 | 0.28497 | 2.7547 | 6.4305 |
| | 3 | 0.1340 | 0.28531 | 2.9126 | 7.2827 |
| | 4 | 0.1329 | 0.26767 | 3.8376 | 7.6173 |
| | 5 | 0.1311 | 0.2597 | 4.3922 | 8.1403 |
| **Slice102** | 2 | 0.1340 | 0.28779 | 2.6866 | 6.6314 |
| | 3 | 0.1344 | 0.2829 | 3.1403 | 7.0547 |
| | 4 | 0.1397 | 0.25795 | 3.5517 | 7.2156 |
| | 5 | 0.1336 | 0.23269 | 4.4888 | 7.8035 |
| **Slice112** | 2 | 0.1375 | 0.28855 | 2.8102 | 6.5551 |
| | 3 | 0.1372 | 0.24958 | 2.9396 | 7.2357 |
| | 4 | 0.1362 | 0.24055 | 4.1515 | 7.5813 |
| | 5 | 0.1349 | 0.22548 | 4.6141 | 7.9576 |

*Table 7. Comparison study of CHPSO and ABF in term of misclassification error in percentage (ME)*

| Image | m | CHPSO | | ABF | |
|---|---|---|---|---|---|
| | | ME (in %) Otsu | ME (in %) Kapur | ME (in %) Otsu | ME (in %) Kapur |
| Slice 022 | 3 | **0.1578** | **1.4977** | 1.98 | 2.04 |
| | 5 | **0.0744** | 1.2437 | 1.4 | **1.14** |
| Slice 032 | 3 | **0.1312** | **1.1307** | 1.82 | 2.18 |
| | 5 | **0.0900** | **0.5876** | 0.85 | 1.01 |
| Slice 042 | 3 | **0.2319** | 3.6760 | 3.24 | **3.05** |
| | 5 | **0.0999** | **0.2194** | 1.1 | 1.27 |
| Slice 052 | 3 | **0.3038** | 11.6213 | 8.03 | **9.16** |
| | 5 | **0.0772** | **0.1719** | 1.33 | 1.40 |
| Slice 062 | 3 | **0.3095** | 9.2941 | 7.47 | **7.73** |
| | 5 | **0.0675** | **0.2305** | 4.33 | 5.57 |
| Slice 072 | 3 | **0.2954** | 9.7454 | 10.98 | **8.01** |
| | 5 | **0.0735** | **0.7059** | 4.97 | 4.31 |
| Slice 082 | 3 | **0.3304** | **11.2282** | 12.27 | 12.33 |
| | 5 | **0.0542** | **0.2791** | 9.02 | 7.24 |
| Slice 092 | 3 | **0.2265** | 11.8704 | 9.49 | **10.29** |
| | 5 | **0.0476** | **0.1634** | 5.53 | 5.97 |
| Slice 102 | 3 | **0.2475** | 7.9017 | 7.25 | **7.63** |
| | 5 | **0.0465** | **0.0231** | 5.49 | 6.98 |
| Slice 112 | 3 | **0.0849** | **0** | 1.77 | 2.38 |
| | 5 | **0.0479** | **0** | 1.11 | 1.59 |

# 5 Conclusion

This paper has presented a new approach for thresholding segmentation of MR brain images using a combination of Convergent Heterogeneous Particle Swarm Optimization (CHPSO) and two classical thresholding techniques. We have shown that utilizing a new strategy of searching named multiswarm can improve the performance of a heuristic algorithm for the problem of image thresholding. In this strategy, the particles are divided into subswarms and each subswarm separately searches the problem space to find the best solution and improve the exploitation while the subswarms have a cooperation with each other and share their information for better exploration to find the global best position and jump from local optimal answers.

Empirical testing on a set of 10 medical images, shows that the proposed approach is significantly robustness with better convergence, in comparison to similar techniques. Instead of reporting classical techniques, we compare our method with the state of the art and powerful heuristic method Amended Bacterial Foraging (ABF). In terms of speed and accuracy also, the CHPSO outperforms the ABF. For visualization of the results, all segmented images have been illustrated. Images show that the details form the original image properly conveyed into the segmented image with more details. In future work, we hope to apply other kinds of multiswarm heuristic algorithms and provide more obvious demonstrations of this technique by applying it to more complex kinds of segmentation and images.

24